\documentclass[lettersize,journal]{IEEEtran} 
\usepackage{amsmath,amsfonts}
\usepackage{algorithm}
\usepackage{algpseudocode}
\usepackage{array}
\usepackage[caption=false,font=normalsize,labelfont=sf,textfont=sf]{subfig}
\usepackage{textcomp}
\usepackage{xcolor}
\usepackage{stfloats}
\usepackage{hyperref}
\usepackage{verbatim}
\usepackage{graphicx}
\usepackage{optidef}
\usepackage[style=ieee,doi=false,isbn=false,url=false,eprint=false]{biblatex}

\newcommand{\R}[1]{\mathbb{R}^{#1}}
\newcommand{\jac}[2]{\frac{\partial { #1  }}{ \partial { #2 }}}
\begin{document}

\title{DiffPills: Differentiable Collision Detection\\ for Capsules and Padded Polygons}

\author{Kevin Tracy$^{1}$, Taylor A. Howell$^{2}$, Zachary Manchester$^{1}$
\thanks{$^1$Kevin Tracy and Zachary Manchester are with The Robotics Institute,  Carnegie Mellon University, Pittsburgh, PA 15213, USA {\tt\small \{ktracy,zacm\}@cmu.edu}}
\thanks{$^{2}$Taylor A. Howell is with the Department of Mechanical Engineering, Stanford University, Stanford, CA 94305, USA {\tt\small thowell@stanford.edu}}
}

\maketitle
\begin{abstract}
Collision detection plays an important role in simulation, control, and learning for robotic systems. However, no existing method is differentiable with respect to the configurations of the objects, greatly limiting the sort of algorithms that can be built on top of collision detection. In this work, we propose a set of differentiable collision detection algorithms between capsules and padded polygons by formulating these problems as differentiable convex quadratic programs. The resulting algorithms are able to return a proximity value indicating if a collision has taken place, as well as the closest points between objects, all of which are differentiable. As a result, they can be used reliably within other gradient-based optimization methods, including trajectory optimization, state estimation, and reinforcement learning methods. 
\end{abstract}
\section{Introduction}
Collision detection algorithms are used to determine if two abstract shapes have an intersection. This problem has been the subject of great interest from the computer graphics and video game communities, where accurate collision detection is a key part of both the simulation as well as the visualization of complex shapes \cite{olvang2010,bergen2004}. Robotics shares a similar interest in collision detection, as it plays a role in both the accurate simulation of systems that make and break contact \cite{howell2022}, as well as a tool for constrained motion planning \cite{howell2019}.   

Popular algorithms for collision detection are the Gilbert, Johnson, and Keerthi (GJK) algorithm \cite{gilbert1988}, its updated variant enhanced-GJK \cite{cameron1997}, and Minkowski Portal Refinement (MPR) \cite{snethen2008,newth2013}. All of these algorithms rely on a set of primitives and corresponding support mappings to calculate either the shortest distance between two objects, or the existence of a collision. While these methods are highly efficient, robust, and mature, there are inherently non-differentiable due to the logic control flow.

While the differentiability of a collision detection algorithm is not as relevant in the computer graphics and video game communities, it is a key enabling technology in robotics. Highly accurate contact physics formulations like that in Dojo \cite{howell2022} rely on differentiable collision detection to simulate realistic contact behavior. This requirement currently limits Dojo to only simple contact interactions between basic primitives like spheres and a floor. In motion planning, collision avoidance constraints are most often formulated with naive spherical keep-out zones \cite{howell2019}.  A differentiable collision detection algorithm enables a variety of more expressive primitives to be utilized in robotic simulation, planning, and learning. 

\begin{figure}[t]
\centerline{\includegraphics[width = 8cm]{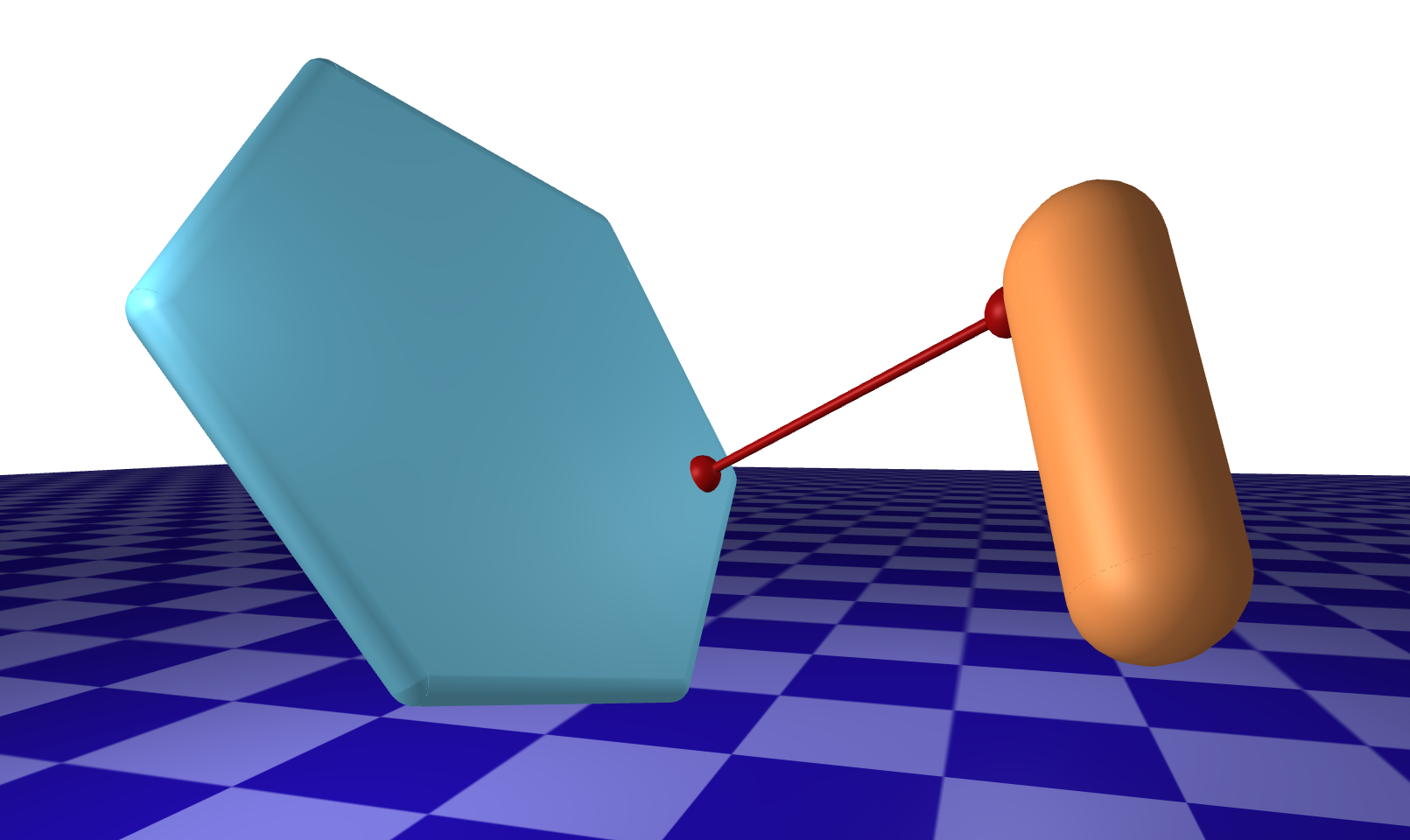}}
\caption{Visualization of a padded polygon (blue) and a capsule (orange). The differential collision detection algorithm computed the proximity value indicating there is no collision, as well as the closest point between the objects in red.  Both of these operations are fully differentiable with respect to the position and orientation of each shape.}
\label{intro_shot}
\end{figure}

In this work, a new approach to collision detection is taken by formulating the routine as a differentiable convex optimization problem. By describing abstract objects as a collection of two primitive types, a capsule and what we call a ``padded" polygon, a variety of more complex geometries can be constructed. For example, arbitrary non-convex geometry can be decomposed into our primitives.  An example of these two primitives is shown in Fig \ref{intro_shot}. The resulting algorithms for collision detection between these primitives work by formulating and solving convex optimization problems that return a continuous proximity value that is positive when there is no collision, and negative when a collision is detected. Using recent advances in differentiable convex optimization, derivatives of this proximity value and the closest points between shapes are calculated with respect to the configurations of the objects. 
\section{Differentiable Convex Optimization}
%

%

In this work, a core part of each collision detection function is the solution to a convex optimization problem. These problems can be solved quickly in polynomial time \cite{boyd2004}, and recent developments in differentiable convex optimization allow for efficient computation of derivatives through these optimization problems \cite{agrawal2019a,amos2019}. 
\subsection{Quadratic Programming}
In computing the collision information for the convex primitives in this work, we will specifically utilize inequality-constrained convex quadratic programs of the following form \cite{boyd2004}:
\begin{mini}
{x}{ \frac{1}{2}x^TPx + c^Tx }{\label{qp}}{}
\addConstraint{Gx}{\leq h.}
\end{mini}
With a primal variable $x \in \R{n}$, quadratic and linear cost terms $P \in \mathbb{S}_+^{n \times n}$ and $c \in \R{n}$, and an inequality constraint described by $G \in \R{l \times n}$ and $h \in \R{l}$. A dual variable $\lambda \in \R{l}$ is introduced for the inequality constraint, and the the Lagrangian for this problem is the following:
\begin{align}
    \mathcal{L}(x,z) &= \frac{1}{2}x^TPx + c^Tx + \lambda^T(Gx-h).
\end{align}
The resulting KKT conditions for stationarity, complementary slackness, primal feasibility, and dual feasibility are:
\begin{align}
    Px + c + G^T\lambda &= 0, \label{kkt:1}\\ 
    D(Gx - h)\lambda &= 0,\label{kkt:2} \\ 
    Gx &\leq h, \label{kkt:3}\\ 
    \lambda &\geq 0, \label{kkt:4}
\end{align}
where $D(\cdot)$ creates a diagonal matrix from the input vector. In this work, a function $x,\lambda = \operatorname{solve\_qp}(P,c,G,h)$ will be the mapping between the description of the problem and the primal and dual solutions to \eqref{qp}.
\subsection{Primal-dual Interior-point Methods} \label{sect:pdip}
Problem \eqref{qp}, can be solved with a primal-dual interior-point method \cite{boyd2004}.  As shown in \cite{mattingley2012}, a primal-dual method with a Mehrotra predictor-corrector \cite{mehrotra1992} uses a variant of Newton's method to iteratively minimize the residuals from \eqref{kkt:1}-\eqref{kkt:2}. 

By introducing a slack variable $s \in \R{l}$ for the inequality constraints, and initializing both $s>0$ and $\lambda >0$, the Newton steps are the following for the affine step:
\begin{align}
    \begin{bmatrix} \Delta x^{\text{aff}} \\ \Delta s^\text{aff} \\ \Delta \lambda^\text{aff} \end{bmatrix} &= K^{-1} \begin{bmatrix} -(Px + c + G^T\lambda) \\ -D(s)\lambda \\ -(Gx + s - h) \end{bmatrix}, \label{pdip:affine}
\end{align}
and for the centering and correcting step:
\begin{align}
    \begin{bmatrix} \Delta x^{\text{cc}} \\ \Delta s^\text{cc} \\ \Delta \lambda^\text{cc} \end{bmatrix} &= K^{-1} \begin{bmatrix} 0 \\ \sigma \mu 1 - D(\Delta s^\text{aff}) \Delta \lambda^\text{aff} \\ 0 \end{bmatrix}, \label{pdip:cc}
\end{align}
where $\sigma \in \R{}$ and $\mu \in \R{}$ are defined in \cite{mattingley2012}, and 
\begin{align}
    K &= \begin{bmatrix} P & 0 & G^T \\ 0 & D(\lambda) & D(s) \\ G & I & 0 \end{bmatrix}.
\end{align}
These two search directions are then added and a line search is used to ensure the positivity of both $s$ and $\lambda$.  Convergence criteria is often based on the norm of the KKT conditions captured in the right-hand side vector in equation \eqref{pdip:affine}. 

There are various methods for solving the linear systems in equations \eqref{pdip:affine} and \eqref{pdip:cc}, and they all exploit the fact that the two linear systems have the same coefficient matrix. This means factorizations only need to take place once and can be re-used in the centering and correcting step computation \cite{nocedal2006, nemirovski2008}. 
\subsection{Differentiating Through a Quadratic Program} \label{diff_a_qp}
At the core of differentiable convex optimization is the implicit function theorem \cite{amos2019}. An implicit function $g:\R{a} \times \R{b} \rightarrow \R{a} $ is defined as:
\begin{align}
    g(y^*,\rho) &= 0 ,\label{ift:res}
\end{align}
for a solution $y^* \in \R{a}$, and problem parameters $\rho \in \R{b}$. By approximating \eqref{ift:res} with a first-order Taylor series, we see 
\begin{align}
    \frac{\partial g}{\partial y} \delta y + \frac{\partial g}{\partial \rho} \delta \rho &= 0 ,
\end{align}
which can be re-arranged to solve for the sensitivities of the solution with respect to the problem parameters:
\begin{align}
    \frac{\partial y}{\partial \rho} &= - \bigg( \frac{\partial g}{\partial y} \bigg)^{-1} \frac{\partial g}{\partial \rho}, \label{eq:ift}
\end{align}
when $\partial g / \partial y$ is invertible.  Problem \eqref{qp} can now be differentiated in a similar fashion by treating the stationarity and complementary slackness optimality conditions in \eqref{kkt:1}-\eqref{kkt:2} as an implicit function of solution variables $x$ and $\lambda$, and problem parameters $P$, $w$, $G$, and $h$. Using this, the sensitivities of the primal and dual variables with respect to the problem parameters can be computed using equation \eqref{eq:ift}.

Instead of naively using the implicit function theorem on \eqref{kkt:1}-\eqref{kkt:2} to calculate the sensitivities, we can leverage the fact that only the left matrix-vector product with another derivative is needed. This means that if we have an arbitrary function $\ell(x): \R{n} \rightarrow \R{}$ and we have the derivative of this function $\jac{\ell}{x}$, we can form the derivatives of $\ell$ with respect to the problem parameters directly using only the primal and dual solutions $x^*$ and $\lambda^*$:
\begin{align}
    \frac{\partial \ell}{ \partial (P)_v} &= \frac{1}{2}(d_xx^{*T} + \lambda^*d_x^T)_v , \label{eq:optnet_P}\\ 
    \frac{\partial \ell}{ \partial (G)_v} &= (D(\lambda^*)d_\lambda x^{*T} + \lambda^*d_x^T)_v, \label{eq:optnet_G}\\ 
        \frac{\partial \ell}{ \partial w} &= d_x, \label{eq:optnet_w}\\ 
    \frac{\partial \ell}{\partial h} &= -D(\lambda^*)d_\lambda, \label{eq:optnet_h}
\end{align}
where $(\cdot)_v$ is the input matrix vectorized, and 
\begin{align}
    \begin{bmatrix} d_x \\ d_\lambda \end{bmatrix} = -\begin{bmatrix} P & G^TD(\lambda^*) \\ G & D(Gx^* - h) \end{bmatrix} \begin{bmatrix} (\frac{\partial \ell }{\partial x})^T \\ 0 \end{bmatrix}. \label{eq:optnet}
\end{align}
The solution to the linear system in \eqref{eq:optnet} can be obtained using the already computed factorization to the primal-dual interior-point steps, meaning all of these derivatives can be computed without any new linear system factorizations \cite{amos2019}.
\section{Capsules} \label{sect:Capsules}
\begin{figure}[htbp]
\centerline{\includegraphics[width = 8cm]{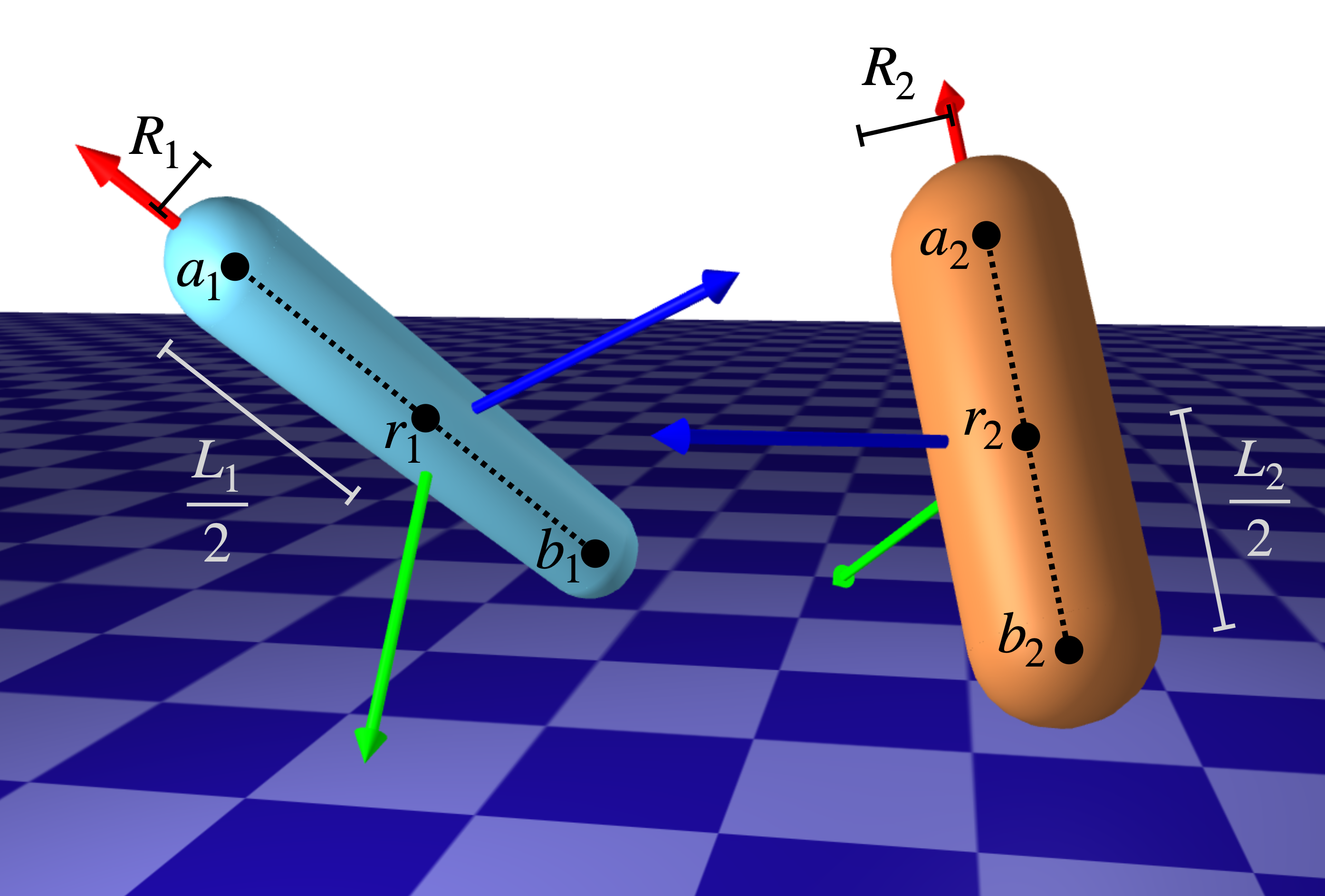}}
\caption{Geometrical description of two capsules as defined by their centroids $r_i$, orthogonal basis $\mathcal{B}_i$, lengths $L_i$, radii $R_i$, and end points $a_i$ and $b_i$. Internal to each capsule is a line segment connecting points $a_i$ and $b_i$.}
\label{two_pills}
\end{figure}

A capsule $i$ in this work will be uniquely described by endpoints $a_i\in \R{3}$ and $b_i\in \R{3}$, and radius $R_i\in \R{}$. In this section, the line segment between points $a_i$ and $b_i$ will be referred to as the central line segment of the capsule, where the line segment can be described by $\theta a_i + (1-\theta)b_i$ for $\theta \in [0,1]$. Mathematically, we can describe the set of points inside of a capsule as any point $x\in \R{3}$ that is within a distance $R_i$ of this central line segment:
\begin{align}
    \{x \,\,\, |\,\,\, \|x - (\theta a_i + (1-\theta)b_i) \| \leq R_i, \,\,\, \exists \, \theta \in [0,1] \}.
\end{align}
\subsection{Collision Detection}
As shown in Fig. \ref{two_pills}, we will now consider two capsules in the same world frame, each described by end points $a_i$, $b_i$, and radii $R_i$. To solve the collision detection problem for these capsules, an optimization problem is formed that solves for the closest points on the two central line segments:
 \begin{mini}
  {p_1,p_2,\theta_1,\theta_2}{ \|p_1 - p_2\|^2 }{\label{capsule_opt}}{}
  \addConstraint{p_1 }{= \theta_1a_1 + (1-\theta_1)b_1 }
  \addConstraint{p_2 }{= \theta_2a_2 + (1-\theta_2)b_2 }
  \addConstraint{0}{\leq \theta_1 \leq 1}{}
  \addConstraint{0}{\leq \theta_2 \leq 1,}{}
 \end{mini}
 where $p_i\in \R{3}$ is constrained to be on the central line segment of capsule $i$, and $\theta_i \in [0,1]$ is introduced to define the line segment in the optimization problem. The resulting quadratic program in \ref{capsule_opt} can be reformulated to eliminate the equality constraints and only solve for $(\theta_1,\theta_2)$ efficiently with either a custom active-set method shown in algorithm \ref{active_set_bound}, or an interior point method as shown in \ref{sect:pdip}. From the solution of this problem, a proximity value $\phi \in \R{}$ will be defined as the following:
 \begin{align}
     \phi = \|p_1 - p_2\|^2 - (R_1 + R_2)^2 \label{caps_prox}
 \end{align}
 which is the squared distance between the closest points on the two central line segments, with the squared sum of the radii subtracted. This results in $\phi > 0$ for no collision, and $\phi \leq 0$ for a collision. 
 
 On top of our newly introduced proximity value, it is also useful to return the closest point in each capsule. The closest point in capsule $i$ to the opposing capsule is denoted $\tilde{p}_i \in \R{3}$, and can be trivially computed given the solution to problem \eqref{capsule_opt}:
 \begin{align}
     \tilde{p}_1 &= p_1 + R_1 \frac{p_2-p_1}{\|p_2-p_1\|} \label{caps_c_1},\\ 
     \tilde{p}_2 &= p_2 + R_2 \frac{p_1-p_2}{\|p_1-p_2\|}  \label{caps_c_2}.
 \end{align}
 \subsection{Implementation}
 When dealing with robotic systems, it is most convenient to express the configuration of a capsule $i$ as a rigid body with an attached reference frame as described with a Cartesian position $r_i \in \R{3}$ and orientation $q_i\in\mathbb{S}^3$ \cite{jackson2021}. From here, a simple function $\operatorname{EndPoints}$ can be constructed that generates the endpoints of the capsule from this rigid body description:
 \begin{align}
     a_i &= r_i + {}^\mathcal{W}Q {}^{\mathcal{C}_i} \begin{bmatrix} L_i/2 & 0 & 0 \end{bmatrix}^T, \\ 
     b_i &= r_i + {}^\mathcal{W}Q {}^{\mathcal{C}_i} \begin{bmatrix} -L_i/2 & 0 & 0 \end{bmatrix}^T,
 \end{align}
 where ${}^{\mathcal{W}}Q {}^{\mathcal{C}_i} \in \R{3 \times 3}$ is the rotation matrix relating the world frame $\mathcal{W}$ to the reference frame on capsule $i$, $\mathcal{C}_i$. 
 
 The equality constraints in the QP \eqref{capsule_opt} can be eliminated by substituting in the values for $p_1$ and $p_2$ in the cost function. The resulting QP can be formulated in the canonical form presented in \eqref{qp} with the following problem data:
 \begin{align}
    P_{c} &= F^TF, & 
    c_{c}  &= F^T (b_1 - b_2), \label{caps_qp_cost} \\ 
    G_{c}  &= \begin{bmatrix} I_2 & -I_2 \end{bmatrix}^T, & 
    h_{c}  &= \begin{bmatrix} 1 & 1 & 0 & 0 \end{bmatrix}^T, \label{caps_qp_cons}
\end{align}
where $F = [(a_1 - b_1), (b_2 - a_2)]$ and the primal variables are $\theta_1$ and $\theta_2$. The resulting algorithm for collision detection between two capsules is outlined in algorithm \ref{alg:capsule}. This algorithm takes in a description of two capsules and returns a proximity value $\phi$ where $\phi>0$ indicates there is no collision. 
\begin{algorithm}
\caption{Capsule{-Capsule} Collision Detection}\label{alg:capsule}
\begin{algorithmic}[1]
\State \textbf{input:} $r_1,q_1,r_2,q_2,R_1,R_2$ 
\State $a_1,b_1,a_2,b_2 \gets \operatorname{EndPoints}(r_1,q_1,r_2,q_2,L_1,L_2)$
\State $P_{c} ,c_{c} ,G_{c} ,h_{c}  \gets$  \eqref{caps_qp_cost} - \eqref{caps_qp_cons}
\State $\theta_1, \theta_2 \gets \operatorname{solve\_qp}(P_{c} ,c_{c} ,G_{c} ,h_{c} )$ 
\State $p_1,p_2 \gets$ recover from \eqref{capsule_opt}
\State $\phi \gets$ \eqref{caps_prox}
\State \textbf{return:} $\phi$
\end{algorithmic}
\end{algorithm}
All of the operations in Algorithm \ref{alg:capsule} are readily differentiable, including the $\operatorname{solve\_qp}$ function as detailed in section \ref{diff_a_qp}. The resulting Jacobians of this proximity value $\phi$ with respect to the position and orientation of the capsule are
\begin{align}
    \frac{d\phi}{dr_i} &= \frac{\partial \phi}{\partial a_i}  + \jac{\phi}{b_i} + \jac{\phi}{(P_c)_v} \jac{(P_c)_v}{r_i} + \jac{\phi}{w_{cc}} \jac{w_{c}}{r_i}, \label{caps_deriv_1}\\
    \frac{d\phi}{dq_i} &= \frac{\partial \phi}{\partial a_i} \frac{\partial a_i}{\partial q_i} + \jac{\phi}{b_i}\jac{b_i}{q_i} + \jac{\phi}{(P_c)_v} \jac{(P_c)_v}{q_i} + \jac{\phi}{w_{c}} \jac{w_{c}}{q_i}, \label{caps_deriv_2}
\end{align}
where $\partial \phi/ \partial (P_c)_v$ and $\partial \phi / \partial w_c$ are both computed using the formulation in \eqref{eq:optnet_P} and \eqref{eq:optnet_w}.
\section{Padded Polygons} \label{sect:polygons}
\begin{figure}[htbp]
\centerline{\includegraphics[width = 8cm]{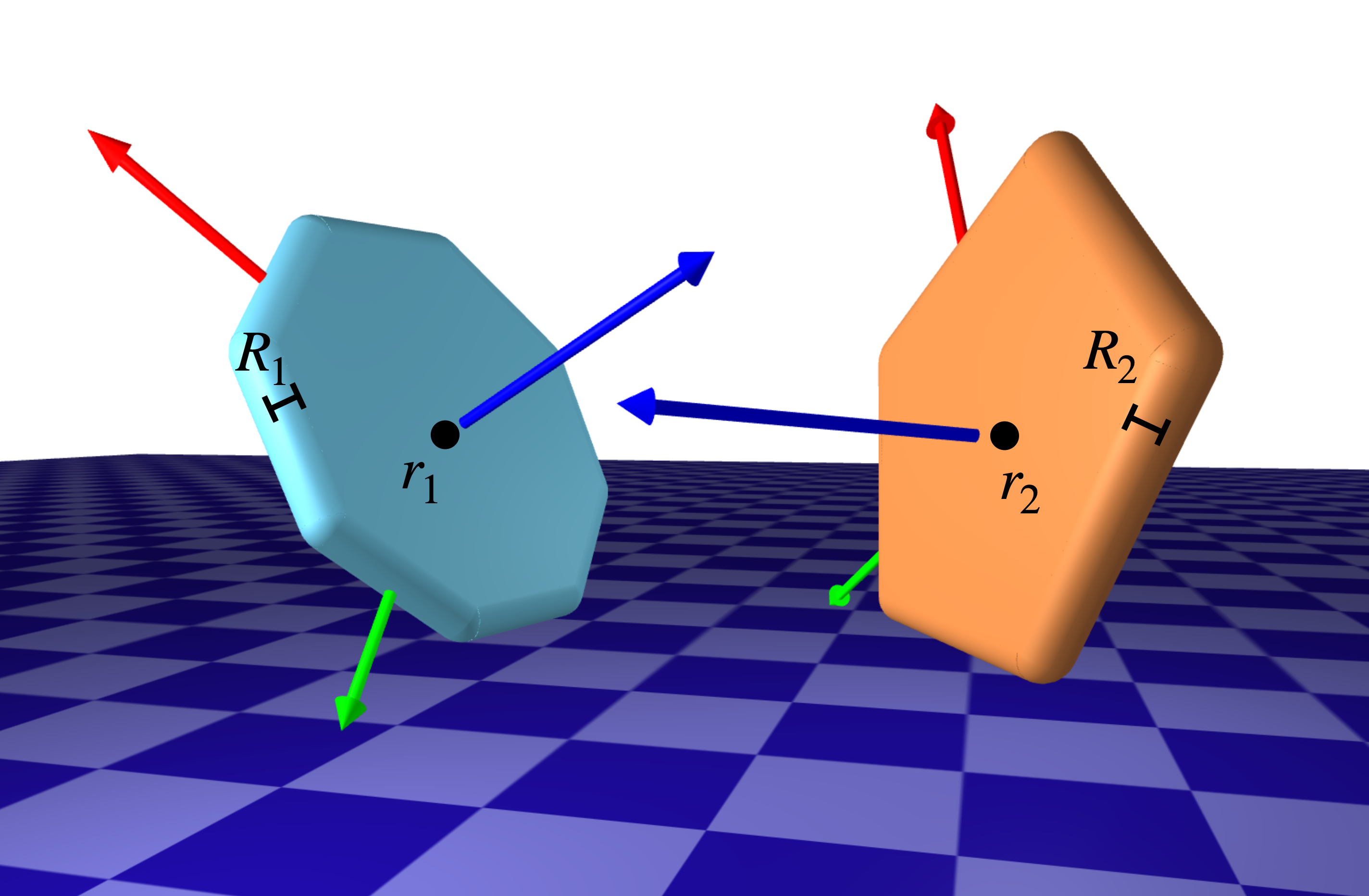}}
\caption{Description of a``padded" polygon, where the shape is defined as the set of points within some radius $R$ of a two-dimensional polygon. A reference frame $\mathcal{B}$ is defined in the center of the polygon, and the polygon exists in the first two basis vectors of $\mathcal{B}$.}
\label{fig:polygon}
\end{figure}
This section will examine a primitive that is defined by a two-dimensional polygon in three-dimensional space with a uniform ``padding" radius of $R_i\in\R{}$ around polygon $i$. In order to represent this mathematically, let us define a polygon in two dimensions with some basis $\mathcal{B}_i$ such that a point $y \in \R{2}$ is within the polygon if $C_iy\leq d_i$.  The origin of the basis $\mathcal{B}_i$ is within the polygon, and the plane that the polygon exists on is spanned by the first two basis vectors of $\mathcal{B}_i$. The origin of basis $\mathcal{B}_i$ is $r_i \in \R{3}$ in the world frame, and ${}^\mathcal{W} Q {}^{\mathcal{B}_i} \in \R{3 \times 3}$ relates this basis to the world frame. 

A point $x \in \R{3}$ is said to be within this padded polygon $i$ if it can be represented as the following:
\begin{align}
    x &= r_i + {}^\mathcal{W} \tilde{Q} {}^{\mathcal{B}_i} y,
\end{align}
where $C_iy\leq d_i$ and ${}^\mathcal{W} \tilde{Q} {}^{\mathcal{B}_i}  \in \R{3 \times 2}$ is the first two columns of ${}^\mathcal{W} Q {}^{\mathcal{B}_i}$. In the rest of this section, $\tilde{Q}_i$ will be shorthand for ${}^\mathcal{W} \tilde{Q} {}^{\mathcal{B}_i}$.
\subsection{Collision Detection}
In a similar fashion to section \ref{sect:Capsules}, the collision detection between two padded polygons will be performed with a convex optimization problem. By specifying the underlying two-dimensional polygons, a quadratic program will solve for the closest points between these polygons. If these points are closer than the sum of the two padding radii, then the two shapes intersect. 

Similar to \eqref{capsule_opt}, we can formulate this optimization problem by introducing variables $p_i \in \R{3}$ that are constrained to be on the two-dimensional polygons, and solving for the closest points on these polygons:
 \begin{mini}
  {p_1,p_2,y_1,y_2}{ \|p_1 - p_2\|^2 }{\label{polygon_opt}}{}
  \addConstraint{p_1 }{=  r_1+ \tilde{Q}_1 y_1 }
  \addConstraint{p_2 }{= r_2+  \tilde{Q}_2 y_2  }
  \addConstraint{C_1y_1}{\leq d_1}{}
  \addConstraint{C_2y_2}{\leq d_2.}{}
 \end{mini}
The proximity value $\phi$ is calculated in the same way as in \eqref{caps_prox}, and the closest points between the two padded polygons can be again computed using \eqref{caps_c_1} - \eqref{caps_c_2}. 
 \subsection{Implementation}
 The variables $p_1$ and $p_2$ and the equality constraints can be eliminated in \eqref{polygon_opt}, and the optimization problem can be reformulated in our inequality-only QP form \eqref{qp}. The primal variables for this new problem are $(y_1, y_2)$, and the following problem data are used:
 \begin{align}
    P_{p} &= F^TF, & 
    c_{p}  &= F^T (r_1 - r_2), \label{polygon_qp_cost} \\ 
    G_{p}  &= \begin{bmatrix} C_1 & 0 \\ 0 & C_2 \end{bmatrix}, & 
    h_{p}  &= \begin{bmatrix} d_1 \\ d_2^T \end{bmatrix}, \label{polygon_qp_cons}
\end{align}
where $F = [\tilde{Q}_1,\, -\tilde{Q}_2]$. With this, the collision detection algorithm for two padded polygons is expressed in algorithm \ref{alg:polygon}.
\begin{algorithm}
\caption{Padded Polygon Collision Detection}\label{alg:polygon}
\begin{algorithmic}[1]
\State \textbf{input:} $r_1,q_1,r_2,q_2,C_1,d_1,C_2,d_2,R_1,R_2$  
\State $P_{p} ,c_{p} ,G_{p} ,h_{p}  \gets$  \eqref{polygon_qp_cost} - \eqref{polygon_qp_cons}
\State $y_1, y_2 \gets \operatorname{solve\_qp}(P_{p} ,c_{p} ,G_{p} ,h_{p} )$ 
\State $p_1,p_2 \gets$ recover from \eqref{polygon_opt}
\State $\phi \gets$ \eqref{caps_prox}
\State \textbf{return:} $\phi$
\end{algorithmic}
\end{algorithm}
This algorithm is made up of entirely differentiable operations and can be differentiated through in the same way as in algorithm \ref{alg:capsule}.  The Jacobians of the proximity value to the state and orientation of each padded polygon is as follows:
\begin{align}
    \frac{d\phi}{dr_i} &= \frac{\partial \phi}{\partial r_i}  + \jac{\phi}{w_p} \jac{w_p}{r_i} \label{polygon_deriv_1}, \\ 
    \frac{d\phi}{dq_i} &= \jac{\phi}{q_i} + \jac{\phi}{(P_p)_v} \jac{(P_p)_v}{q_i} + \jac{\phi}{w_p} \jac{w_p}{q_i}, \label{polygon_deriv_2}
\end{align}
where once again $\partial \phi/ \partial (P_p)_v$ and $\partial \phi / \partial w_p$ are both computed using the formulation from \eqref{eq:optnet_P} and \eqref{eq:optnet_w}.  It's important to note that $\phi$ is a function of $p_1$ and $p_2$, which are different for the padded polygon than it is for the capsule.  To calculate the derivatives of $\phi$, simply substitute in the values of $p_1$ and $p_2$ from \eqref{polygon_opt} into \eqref{caps_prox} and differentiate. 
\subsection{Polygon and Capsule Detection}
Collisions between a padded polygon and a capsule can be computed in the same way \eqref{capsule_opt} and \eqref{polygon_opt}. The main idea is the same in that we are finding the closest points between the central line segment of the capsule, and the two-dimensional polygon. If the distance between these two closest points is greater than the sum of the radii, there is no collision. This optimization problem is as follows:
 \begin{mini}
  {p_1,p_2,\theta_1,y_2}{ \|p_1 - p_2\|^2 }{\label{polygon_caps_opt}}{}
  \addConstraint{p_1 }{=  \theta_1a_1 + (1-\theta_1)b_1 }
  \addConstraint{p_2 }{= r_2+  \tilde{Q}_2 y_2  }
  \addConstraint{0}{\leq \theta_1 \leq 1}{}
  \addConstraint{C_2y_2}{\leq d_2,}{}
 \end{mini}
where again $p_1$ and $p_2$ can be eliminated resulting in an inequality-only QP that is solving for $[\theta_1, y^T]^T$. The problem data for this QP are the following:
 \begin{align}
    P_{cp} &= F^TF, & 
    c_{cp}  &= F^T (b_1 - r_2), \label{polygon_caps_qp_cost} \\ 
    G_{cp}  &= \begin{bmatrix} D([1,-1]) & 0 \\ 0 & C_2 \end{bmatrix}, & 
    h_{cp}  &= \begin{bmatrix} 1 & 0 & d_2^T \end{bmatrix}^T, \label{polygon_caps_qp_cons}
\end{align}
where $F = [(a_1 - b_1), -\tilde{Q}_2]$. The algorithm for collision detection between these two primitives is detailed in Algorithm \ref{alg:polygon_capsule}, where again it is made up of entirely differentiable operations. The Jacobians of the proximity value with respect to the position and orientation of the capsule are calculated with equations \eqref{caps_deriv_1}-\eqref{caps_deriv_2}, and the Jacobians of the proximity value with respect to the padded polygon's position and orientation are calculated with equations \eqref{polygon_deriv_1}-\eqref{polygon_deriv_2}. 
\begin{algorithm}
\caption{Padded Polygon and Capsule Collision Detection}\label{alg:polygon_capsule}
\begin{algorithmic}[1]
\State \textbf{input:} $r_1,q_1,r_2,q_2,a_1,b_1,C_2,d_2,R_1,R_2$  
\State $P_{cp} ,c_{cp} ,G_{cp} ,h_{cp}  \gets$  \eqref{polygon_caps_qp_cost} - \eqref{polygon_caps_qp_cons}
\State $\theta_1, y_2 \gets \operatorname{solve\_qp}(P_{cp} ,c_{cp} ,G_{cp} ,h_{cp} )$ 
\State $p_1,p_2 \gets$ recover from \eqref{polygon_caps_opt}
\State $\phi \gets$ \eqref{caps_prox}
\State \textbf{return:} $\phi$
\end{algorithmic}
\end{algorithm}
\section{Motion Planning Example}
 To demonstrate the utility of differentiable collision detection between these primitives, the formulation presented in section \ref{sect:Capsules} will be used for a motion planning problem. Two cars are modeled as capsules, with one larger car being stationary, and another smaller car being controlled with acceleration and steering-angle rate commands. The equations of motion for this simple car are as follows:
 \begin{align}
     \dot{p}_x &= v \cos(\theta), &  \dot{v} &= u_1, \\ 
     \dot{p}_y &= v \sin(\theta), &  \dot{\gamma} &= u_2,
 \end{align}
 where $p_x,p_y \in \R{}$ is the position of the car, $v \in \R{}$ is the velocity, $\gamma \in \R{}$ is the steering angle, and $u \in \R{2}$ are the control inputs. A trajectory optimization problem is formed where the objective is encouraging the controlled car to hit a desired configuration, but the stationary car is directly in the way. To incorporate this collision avoidance constraint into the problem, the differential collision detection algorithm \ref{alg:capsule} was used where the proximity value was constrained to be $\phi \geq 0$. 
 
\begin{figure}[htbp]
\centerline{\includegraphics[width = 8cm]{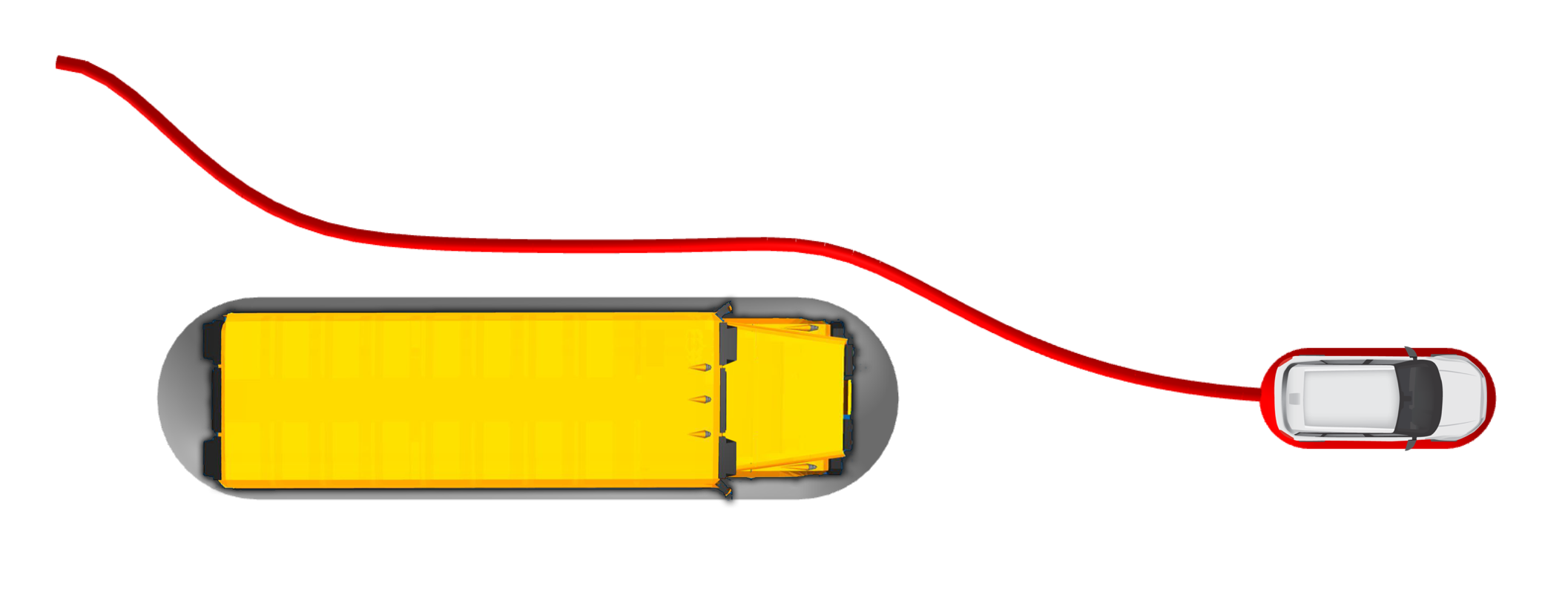}}
\caption{Trajectory optimization for a car with a collision avoidance constraint. The car starts on the left side and is trying to get to the lower right side, but it must avoid colliding with the yellow school bus.  By explicitly specifying this collision avoidance constraint with our proposed differential collision detection algorithm, a trajectory optimizer is able to converge on a feasible and optimal solution.}
\label{fig:cars}
\end{figure}

 The trajectory optimizer ALTRO \cite{howell2019} was used for this problem, where derivatives of the collision avoidance constraint were required. As shown in Fig. \ref{fig:cars}, the optimal trajectory from ALTRO shows the car starting on the left and traveling to the right towards the goal while avoiding the stationary car. 
 
\subsection{Open Source Implementation}
An open source implementation of these algorithms in Julia \cite{bezanson2017} is available at  \url{https://github.com/kevin-tracy/DiffPills.jl}.   
\printbibliography

\vspace{12pt}

\section*{Appendix A}
\section*{Active Set QP Solver}\label{sect:active_set}
For quadratic programs where the only constraints are inequality bound constraints on two primal variables $x\in\R{2}$, an active set method is able to outperform primal-dual interior-point methods. This optimization problem looks like the following:
 \begin{mini}
  {x}{ \frac{1}{2}x^TPx + c^Tx }{\label{bound_qp}}{}
\addConstraint{\begin{bmatrix} I \\ -I \end{bmatrix} x}{\leq \begin{bmatrix} 1 \\ 0 \end{bmatrix} }
 \end{mini}
And since the feasible set is a two-dimensional box, the solution must be either in the middle of the box, one of the four corners, or on one of the four sides of the box. The active-set method shown in algorithm \ref{active_set_bound} calculates these 9 points and returns the feasible one with minimum cost. The dual variables $\lambda \in \R{4}$ are then backed out using algorithm \ref{active_set_duals}.
\begin{algorithm}
\caption{Two-dimensional Active Set QP Solver}\label{active_set_bound}
\begin{algorithmic}[1]
\State \textbf{function} $\operatorname{active\_set\_2D}(P,c)$ 
\State $x^* \gets -P^{-1}c $  \Comment{compute unconstrained solution}
\If{$0\leq x^* \leq 1$} \Comment{return solution if feasible}
\State $\lambda \gets 0$
\Else{} \Comment{generate 8 candidate solutions}
    \State $x^{(1)} = [1, -(P_2 + c_2)/P_3]$
    \State $x^{(2)} = [0,-c_2/P_3]$
    \State $x^{(3)} = [-(P_2 + c_1)/P_1, 1]$
    \State $x^{(4)} = [-c_1/P_1, 0]$
    \State $x^{(5)} = [0,0]$
    \State $x^{(6)} = [0, 1]$
    \State $x^{(7)} = [1,0]$
    \State $x^{(8)} = [1, 1]$
    \State $x^* \gets$ feasible $x^{(i)}$ with minimum cost
    \State $\lambda \gets \operatorname{recover\_duals}(x^*,P,c)$
\EndIf
\State \textbf{return} $x^*,\, \lambda$ 
\end{algorithmic}
\end{algorithm}
\begin{algorithm}
\caption{Recover Duals from Primal Solution}\label{active_set_duals}
\begin{algorithmic}[1]
\State \textbf{function} $\operatorname{recover\_duals}(x^*,P,c)$ 
\State $y \gets -Px^* - c$
\State $\lambda \gets 0$
\State $\epsilon \gets 1\cdot 10^{-12}$
\If{$y_1\geq \epsilon$}
\State $\lambda_1 \gets y_1$
\EndIf
\If{$y_2\geq \epsilon$}
\State $\lambda_2 \gets y_2$
\EndIf
\If{$y_1\leq -\epsilon$}
\State $\lambda_3 \gets -y_1$
\EndIf
\If{$y_2\leq -\epsilon$}
\State $\lambda_4 \gets -y_2$
\EndIf
\State \textbf{return} $\lambda$ 
\end{algorithmic}
\end{algorithm}

\section*{Appendix B}
\section*{Interior-Point Linear System Solver}\label{sect:pdip_ls}
In our primal-dual interior-point method, we solve two linear systems (\eqref{pdip:affine} and \eqref{pdip:cc}) to compute the Newton steps. This linear system takes the following form for an arbitrary right hand side vector designated by $v_1,v_2,v_3$.
\begin{align}
    \begin{bmatrix} P & 0 & G^T \\ 0 & D(\lambda) & D(s) \\ G & I & 0 \end{bmatrix} \begin{bmatrix} \Delta x \\ \Delta s \\ \Delta \lambda \end{bmatrix} &= \begin{bmatrix} v_1 \\ v_2 \\ v_3 \end{bmatrix},
\end{align}
The resulting algorithm for solving this linear system via a single Cholesky decomposition is shown in \ref{pdip_ls}, where it is important to note that the Cholesky factorization only needs to take place during the first solving of \eqref{pdip:affine}, and can be cached and reused for the subsequent solve of \eqref{pdip:cc}.
\begin{algorithm}
\caption{Interior-Point Linear System Solver}\label{pdip_ls}
\begin{algorithmic}[1]
\State \textbf{function} $\operatorname{solve\_pdip\_linear\_system}(P,G,\lambda,s,v_1,v_2,v_3)$ 
\State $W \gets D(\lambda / s)$ 
\State $L \gets \operatorname{cholesky}(P + G^TWG)$
\State $\Delta x \gets L^{-T}L^{-1}(-v_1 + G^TW(-v_3 + v_2/\lambda))$
\State $\Delta s \gets -G\Delta x - v_3$
\State $\Delta \lambda \gets -(v_2 + (z \circ \Delta s)) / s$
\State \textbf{return:} $\Delta x, \Delta s, \Delta \lambda$
\end{algorithmic}
\end{algorithm}

\end{document}